\documentclass[a4paper]{article}

\usepackage{INTERSPEECH2019}
\usepackage{amsmath, hyperref}
\usepackage{algorithm}
\usepackage{algorithmic}

\title{A New GAN-based End-to-End TTS Training Algorithm}
\name{Haohan Guo$^*$, Frank K. Soong$^\dag$, Lei He$^\dag$, Lei Xie$^*$\thanks{Work performed as an intern at Microsoft.}}
\address{$^*$School of Computer Science, Northwestern Polytechnical University, Xi’an, China \\
  $^\dag$Microsoft AI \& Research, Beijing, China}
\email{\{hhguo,lxie\}@nwpu-aslp.org, \{frankkps, helei\}@microsoft.com}

\begin{document}

\maketitle
\begin{abstract}
End-to-end, autoregressive model-based TTS has shown significant performance improvements over the conventional one. However, the autoregressive module training is affected by the “exposure bias”, or the mismatch between the different distributions of real and predicted data. While real data is available in training, but in testing, only predicted data is available to feed the autoregressive module. By introducing both real and generated data sequences in training, we can alleviate the effects of the exposure bias. We propose to use Generative Adversarial Network (GAN) along with the key idea of “Professor Forcing” in training. A discriminator in GAN is jointly trained to equalize the difference between real and predicted data. In AB subjective listening test, the results show that the new approach is preferred over the standard transfer learning with a CMOS improvement of 0.1. Sentence level intelligibility tests show significant improvement in a pathological test set. The GAN-trained new model is also more stable than the baseline to produce better alignments for the Tacotron output.
\end{abstract}

\section{Introduction}

Statistical parametric text-to-speech (TTS) is a sequence generator, which generates a sequence of speech samples according to the input text or phoneme sequence. To achieve better intelligibility, naturalness and expressiveness, enhancing the model's prediction capability is very important. From HMM \cite{tokuda2000speech} and DNN \cite{ze2013statistical} to LSTM \cite{zen2015acoustic} and BLSTM \cite{fan2014tts}, effective sequence modelling plays an important role in TTS. In recent years, autoregressive (AR) model has been widely used in sequence-to-sequence model to further improve the performance of a sequential model. It has shown significant improvement in speech synthesis, such as autoregressive acoustic model \cite{wang2018autoregressive, wang2017autoregressive}, WaveNet-based \cite{van2016wavenet} or WaveRNN-based \cite{kalchbrenner2018efficient} neural vocoder, and end-to-end TTS system \cite{wang2017tacotron, shen2018natural, ping2018clarinet}.

\begin{figure}[htp]
  \centering
  \includegraphics[width=8cm]{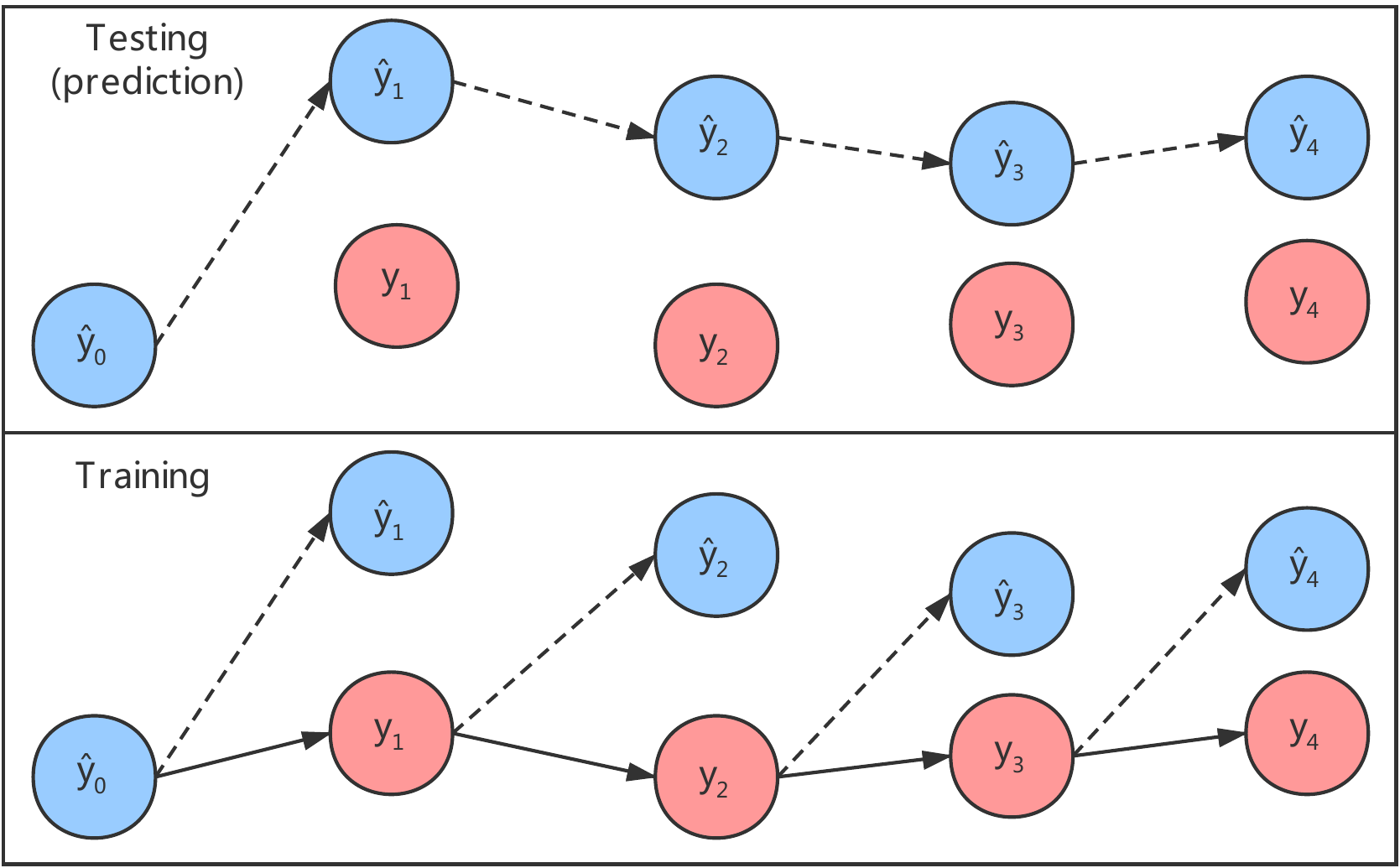}
  \caption{The difference between training (teacher forcing) and testing (prediction) of AR model ($\hat y$: predicted, $y$: real)}
  \label{fig:exposure_bias}
\end{figure}

Autoregressive model specifies that the output sample $\hat{y}_t$ depends on its own previous samples $\hat{y}_{1:t-1}$, which is written as:
\begin{equation}
    p(\hat{y}_{1:T}|X, \Theta) = \prod_{t=1}^{T} p(\hat{y}_t|\hat{y}_{1:t-1}, X, \Theta)
\label{con:autoregressive}
\end{equation}
Here, $X$ and $\Theta$ denote the inputs and the network weights. Although AR model has improved an end-to-end TTS model, the conventional training algorithm (also known as \textit{teacher forcing} \cite{williams1989learning}) has a intrinsic problem in training, named \textit{exposure bias} \cite{ranzato2015sequence}. As shown in Fig.\ref{fig:exposure_bias}, in training, the model is only exposed to real data, which predicts output $\hat{y}_t$ given the real data of previous samples as input, written as
\begin{equation}
    p(\hat{y}_{1:T}|X, \Theta) = \prod_{t=1}^{T} p(\hat{y_t}|y_{1:t-1}, X, \Theta)
\label{con:teacher_forcing}
\end{equation}
But the model can only predict the next step using its own predicted samples in testing. The model distribution, however, can not be the same as the real one, so the discrepancy between these two distributions can quickly accumulate errors in decoding. In the end-to-end TTS system (e.g. \textit{Tacotron}), we often adopt a \textit{data dropout} strategy to alleviate the above problems, which randomly discards part of feedback information in both training and testing to reduce the autoregressive effect in prediction such the generation can rely more on the linguistic information which is available in both training and testing. But it is still not enough to avoid the exposure bias, especially in decoding a long sequence.

\begin{figure*}[!htp]
    \centering
    \includegraphics[width=17cm]{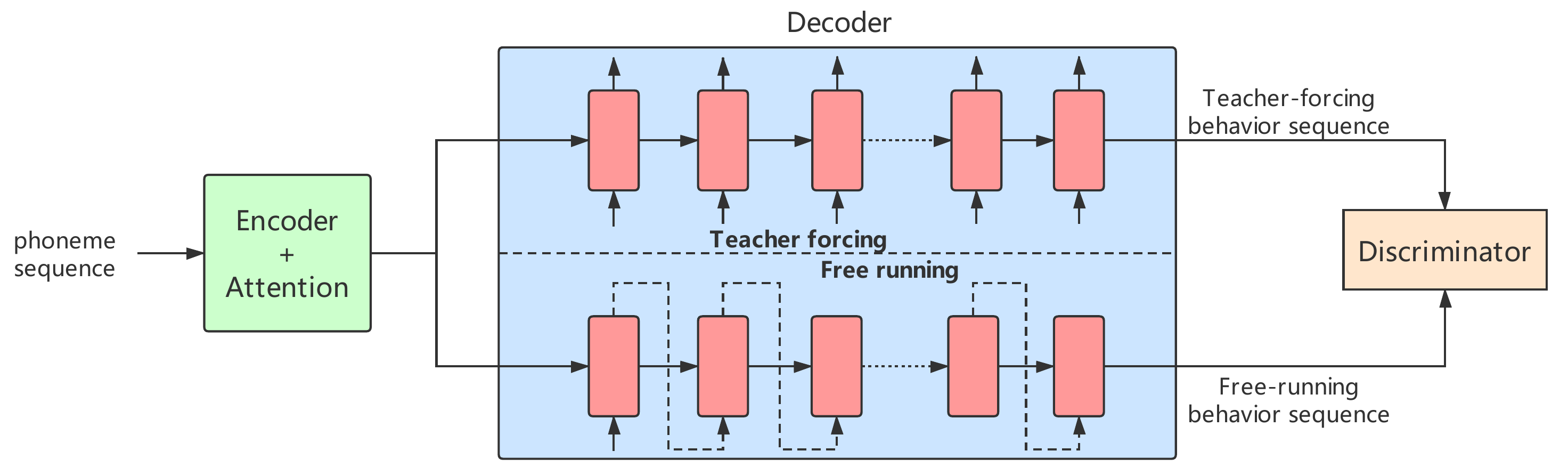}
    \caption{The framework of GAN-based end-to-end TTS training algorithm}
    \label{fig:train_generator}
\end{figure*}

Exposure bias is fundamentally caused by the mismatch between autoregressive predictions and real data used in training. So, we can avoid it by using predicted samples in training the autoregressive model. The widely used algorithm is to feedback the generated data in training with a sampling strategy, e.g. \textit{data as demonstrator (DAD)} \cite{venkatraman2015improving} and \textit{scheduled sampling (SS)} \cite{bengio2015scheduled}. For example, in training with scheduled sampling, we decide whether feedback real data according with a certain probability in each time step, or we will feedback the generated data. The probability will decrease based upon an annealing schedule. Since this training algorithm ignores the temporal dependency of the sequence \cite{huszar2015not}, it may result in misalignment between the target and the predicted sequence. Thus it forces the model trained with MSE to predict possibly an incorrect sequence (see 3.1.2 in \cite{ranzato2015sequence}). How to introduce the complete generated sequence properly in training is necessary for avoiding the exposure bias in the autoregressive model.

Recently adversarial training has been used in many sequential training tasks, e.g. text classification\cite{miyato2016adversarial}, machine translation\cite{wu2017adversarial}, speech recognition\cite{sun2018domain}, etc. In domain adaptation \cite{ganin2016domain}, it has been successfully applied to help learning a domain-invariant representation to improve predictions and model generalization in the target domain. \textit{Professor Forcing} \cite{lamb2016professor} describes a GAN-based adversarial training for generative autoregressive model, which can make predictions with features that cannot be discriminated between real and model distributions. Inspired by it, we propose a new GAN-based, end-to-end TTS training algorithm to introduce generated sequence in training to avoid exposure bias in the autoregressive decoder.

In this paper, we will introduce \textit{Professor Forcing} first, then present our proposed training framework and its training algorithm. Finally, we compare the performance of the training algorithms in different aspects with two subjective evaluation methods. The experimental results show that GAN-based training algorithm can significantly improve the model in different aspects, including naturalness and generalization. We compared it with scheduled sampling to show it is more effective and proper for end-to-end TTS training.

\section{Methods}

\subsection{Professor Forcing}

There are two modules in \textit{Professor Forcing}, a generative RNN (generator) and a discriminator. To introduce the complete predicted sequence in training, the generator will generate sequences in two different modes, teacher forcing (TF) and free running (FR, iteratively generate the sequential predictions). Discriminator is trained as a probabilistic classifier to determine in which mode the behavior sequence $b$ (chosen hidden states and output values) is generated.

The training process of \textit{Professor Forcing} is different from teacher forcing. There are two training objectives for the generator. The first one is to maximize the likelihood of data (depending on the task) using the output sequence generated in the teacher forcing mode. The second one is to equalize the discriminator so as to force the distributions of hidden states to be close to each other. This adversarial process reduces the discrepancy between real and model distributions of the model.

When the end-to-end TTS tries to synthesize a long sentence, it becomes vulnerable to the error accumulation in the AR process. Inspired by \textit{Professor Forcing}, we propose a GAN-based end-to-end TTS training algorithm to train a better autoregressive decoder by avoiding exposure bias.

\subsection{GAN-based End-to-end TTS Training Algorithm}

There have been some studies on GAN in TTS in the past two years, such as GAN-based post filter \cite{Kaneko_2017_Interspeech}, and GAN-based multi-task for TTS \cite{yang2017statistical, saito2018statistical}. These algorithms focus on the generated output sequence of acoustic model, trying to make the outputs be closer to the real data. Our proposed algorithm is different. Specifically, our algorithm focuses on the hidden states of the autoregressive decoder in the end-to-end TTS model, trying to make the behavior sequences generated in different modes be similar to each other. The proposed algorithm is introduced two parts, one is the training framework and model structure, the other is the training algorithm.

\subsubsection{Training Framework \& Model Structure}

As shown in Fig.\ref{fig:train_generator}, we have two models, which are the original end-to-end TTS model as generator, and a discriminator. In this paper, we adopt \textit{Tacotron2} \cite{shen2018natural} as the generator, which has shown good performance in generating high-quality speech. The model structure of discriminator in \textit{Professor forcing} is too simple, which can easily lead to training failure or no convergence. To meet our requirements for stable and effective training, we propose a new model structure based on \textit{Self-Attention GAN (SAGAN)} \cite{zhang2018self} in the discriminator.

\begin{figure}[htp]
  \centering
  \includegraphics[width=8cm]{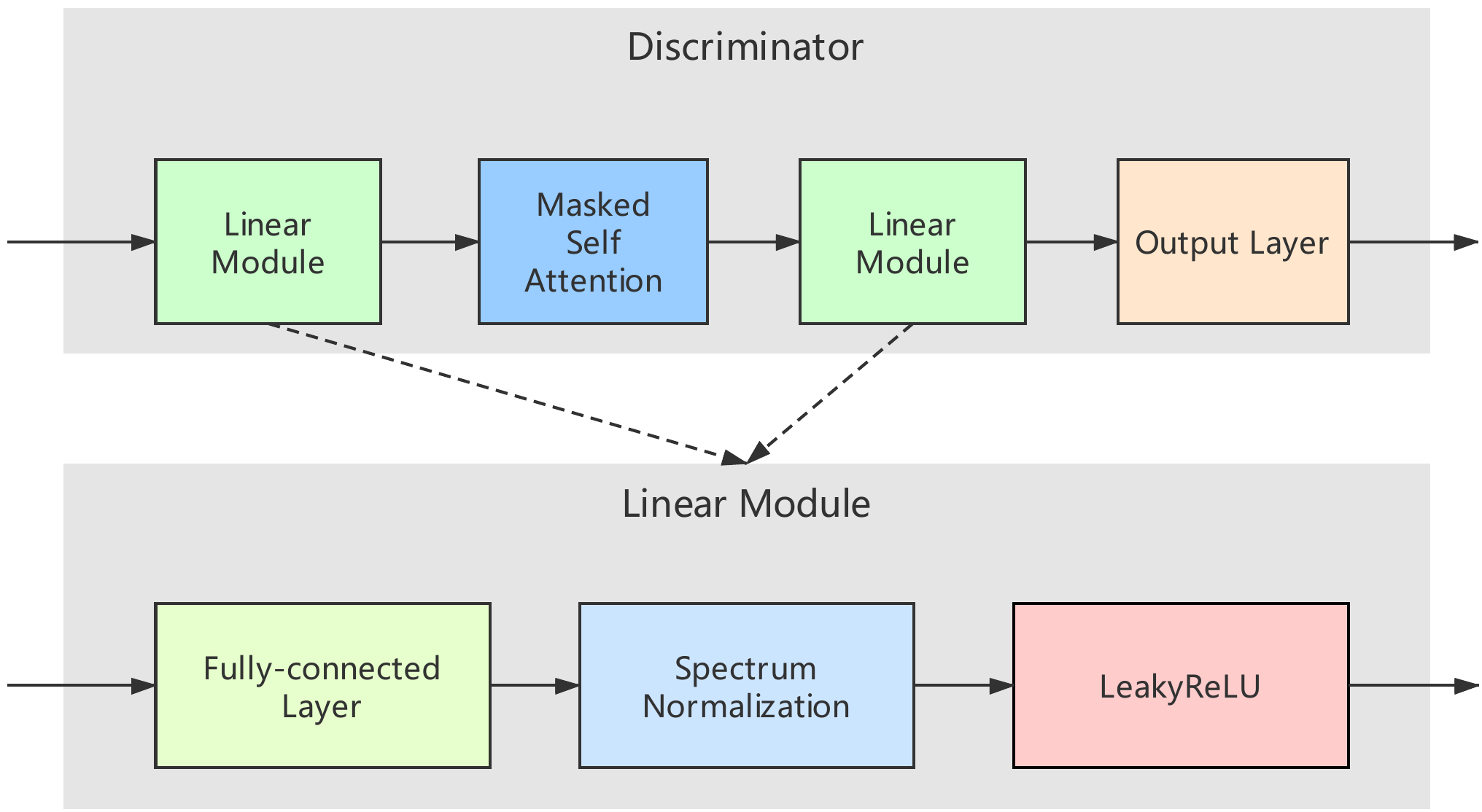}
  \caption{Model architecture of the discriminator}
  \label{fig:discriminator}
  \vspace{-0.3cm}
\end{figure}

Fig.\ref{fig:discriminator} shows the model structure of the discriminator, which has two main components: a linear module and masked self attention. Linear module is composed of a fully connected layer, spectrum normalization and an activation layer (leaky ReLU). Spectrum normalization \cite{miyato2018spectral} can help control Lipschitz constant to stabilize the training of the discriminator and back propagate more effective gradients to the generator, in case of mode collapse and a non-converging generator. Self-attention has shown excellent performance on modelling long-range dependency in many tasks, including: image generation \cite{zhang2018self}, machine translation \cite{vaswani2017attention} and TTS \cite{li2018close}. Because autoregressive decoder is a uni-directional model, the cumulative errors are caused by its history. So we adopt masked self-attention in this work to make the discriminator focus on the impact of the history.

\subsubsection{Training Algorithm}

The discriminator is trained to distinguish in which mode the behavior sequence is generated, either teacher forcing or free running. But the training criterion is different. We train it by minimizing the hinge version of the adversarial loss, which has shown good performance in GAN-based image generation.
\begin{equation}
\begin{split}
    L_D = &-\mathbb{E}_{(x,y) \sim data}[min(0, -1 + D(B_t(x, y)))] \\
          &- \mathbb{E}_{(x,y) \sim data}[min(0, -1 - D(B_f(x)))]
\label{con:teacher_forcing}
\end{split}
\end{equation}
$x, D(*)$ refer to the input sequence and the classification results of the discriminator. We adopt the sequence which is composed of the hidden states of the attention RNN layer and the decoder RNN layer in Tacotron2 as the behavior sequence $b$ to the discriminator. $B_t(*)$ and $B_f(*)$ refer to the behavior sequences generated in teacher forcing mode and free running mode, respectively.

The generator (TTS model) has two tasks in the framework. The first one is to minimize the loss $L_T$ in Tacotron2 between the sequence generated in teacher forcing mode and the target sequence. The second task is to fool the discriminator by making the teacher-forcing and the free-running behavior sequences to be similar, distributions wise. We add a weighting coefficient to balance the two losses for more stabilised training. The training criterion of the generator is defined as
\begin{equation}
\begin{split}
    &L_{G} = L_T - \alpha * (D(B_f(x)) - D(B_t(x, y)))
\end{split}
\label{con:generator}
\end{equation}
Eq.\ref{con:generator} can be considered as a regularizer to restrain the model from over-fitting the distribution of teacher forcing in the training stage.

The training process has two phases: 1. we pre-train an end-to-end TTS model in the teacher forcing mode; 2. we train the TTS model and discriminator in turn. When the discriminator performance is below a lower bound, we will not back propagate the ``bad'' gradients from discriminator to update the generator parameters. Also, we will clamp the discriminator performance to a preset upper bound so as to prevent the discriminator from being too good to continue the training process. So we often test the accuracy every hundreds of steps. The detailed training algorithm is shown below.

\captionsetup[algorithm]{labelformat=empty}
\begin{algorithm}[htb]
\floatname{algorithm}{}
\caption{\textbf{GAN-based end-to-end TTS training algorithm}}
  \label{alg:Framwork}
  \begin{algorithmic}[1]
    \REQUIRE  ~~\\
      Training set: $D = \{x_k, y_k\}_{k=1}^{K}$\\
      \quad $x_k$: phoneme sequence, $y_k$: acoustic feature sequence \\
      Steps for pre-training and GAN-based training: $N_p$, $N_g$ \\
      The required range of the accuracy: [$R_L, R_U$] \\
      The period of testing discriminator accuracy: $N_s$
    \ENSURE ~~\\
      $\theta_g$: TTS model \\
    \STATE Initialize TTS model $\theta_g$, discriminator $\theta_d$
    \STATE Initialize states $s_g = False$, $s_d = True$ 
    \STATE Pre-train $\theta_g$ in teacher forcing mode for $N_p$ steps.
    \FOR{$i = 0; i < N_g; i = i + 1$}
        \STATE Read a batch from $D$, and decode it in two modes
        \STATE Update $\theta_g$ \\
        \quad\textbf{if} $s_g == False$ \\ \quad\quad Back propagate the gradient of $L_T$, update $\theta_g$ \\
        \quad\textbf{else} \\ \quad\quad Back propagate the gradient of $L_G$, update $\theta_g$
        \STATE Update $\theta_d$ \\
        \quad\textbf{if} $s_d == True$ \\ \quad\quad Back propagate the gradient of $L_D$, update $\theta_d$\\
        \STATE \textbf{if} $i$ mod $N_s$ == $0$, update $s_g, s_d$ \\
        \quad Get $accuracy$ of the discriminator on the training set \\
        \quad \textbf{if} $accuracy > R_L$, $s_g = True$; \textbf{else}, $s_g = False$ \\
        \quad \textbf{if} $accuracy < R_U$, $s_d = True$; \textbf{else}, $s_d = False$
    \ENDFOR
    \RETURN $\theta_g$
    \end{algorithmic}

\end{algorithm}

\section{Experiments}

\subsection{Training Setup}

We use Tacotron2 \cite{shen2018natural} as TTS model, include WaveNet as vocoder for all experiments. We use one-hot feature as input, which contains phonemes, punctuation and the blank between two adjacent words. The model output is an 80-channel Mel spectrum (12.5 ms frame shift, 50 ms frame length), one frame at a time. The model structure of the discriminator has been shown in Fig.\ref{fig:discriminator}, which has 1536-dim input, 512-dim hidden size, and 1-dim output.

When we calculate $L_T$ in GAN-based algorithm, teacher forcing can also be replaced with scheduled sampling to generate sequence. We train four TTS models with 4 different training algorithms: teacher forcing (TF), scheduled sampling (SS), GAN-based algorithm with teacher forcing (TF-GAN) and GAN-based algorithm with scheduled sampling (SS-GAN, replace TF in TF-GAN with SS). These experiments are performed based on an American English speech data set, which has 14 hours of speech, recorded by a single female speaker.

All models are trained with a batch size of 128 sequences. We train these models using the Adam optimizer with $\beta_1=0.9$, $\beta_2=0.999$. The learning rate is exponentially decayed from $10^{-3}$ to $10^{-5}$ after 50,000 iterations. The TF model trained with 100,000 steps is set to be the baseline model. In SS, TF-GAN and SS-GAN training, we adopt the TF model trained with 50,000 steps as the pre-trained model, and train it for another 50,000 steps with these algorithms. The scheduled sampling strategy is to use real data with a linear decay, from probability 1 to 0.5, in the first 50,000 steps. We set the initial learning rate $lr_g = 10^{-3}, lr_d = 10^{-3}$, adversarial weight $\alpha = 10^{-3}$ for GAN-based algorithms. The range of the required discriminator accuracy is set to $75\% \sim 97\%$.

\subsection{Subjective Evaluation}

We design two TTS test sets to compare these algorithms in two aspects, speech quality and model generalization (stability). We use the common test set, which contains 50 typical sentences used in news and general conversation, to compare the performance of these models in speech quality and naturalness by a CMOS test. Each pair of samples is rated by 10 native English speakers on a scale from -3 to 3 with 1 point discrete increments. Another test set containing $225$ sentences is used to evaluate generalization of these models with an intelligibility test. These sentences have richer text content, such as long sentence, URL, the sequence of numbers or characters, abbreviation, etc. The test sentences and their contextual information are not well covered in the training set, so that the audios synthesized by these sentences tend to have lower intelligibility. We use it to evaluate the generalization capability of these models by the corresponding diagnostic sentence level intelligibility tests. The listeners need to mark a sentence unintelligible when any part of it is unintelligible in listening.\footnote{Samples are available at \url{https://hhguo.github.io/demo/publications/GANTTS/index.html}}

\begin{table}[htp]
\centering
\setlength{\belowcaptionskip}{5pt}
\begin{tabular}{ccccccccc}
\hline
System B & CMOS & \multicolumn{3}{c}{Preference (\%)} \\ \cline{3-5}
 &  (p-value)  & TF & Neutral & System B \\ \hline
SS & -0.04 (0.22) & 42.80 & 16.60 & 40.60 \\ \hline
TF-GAN & 0.10 (0.02) & 22.73 & 49.21 & 28.06 \\ \hline
SS-GAN & 0.01 (0.40) & 28.60 & 39.00 & 32.40 \\ \hline
\end{tabular}
\caption{The results of the CMOS tests}
\label{tab:cmos}
\vspace{-0.8cm}
\end{table}

\begin{figure}[htp]
  \centering
  \includegraphics[width=8cm]{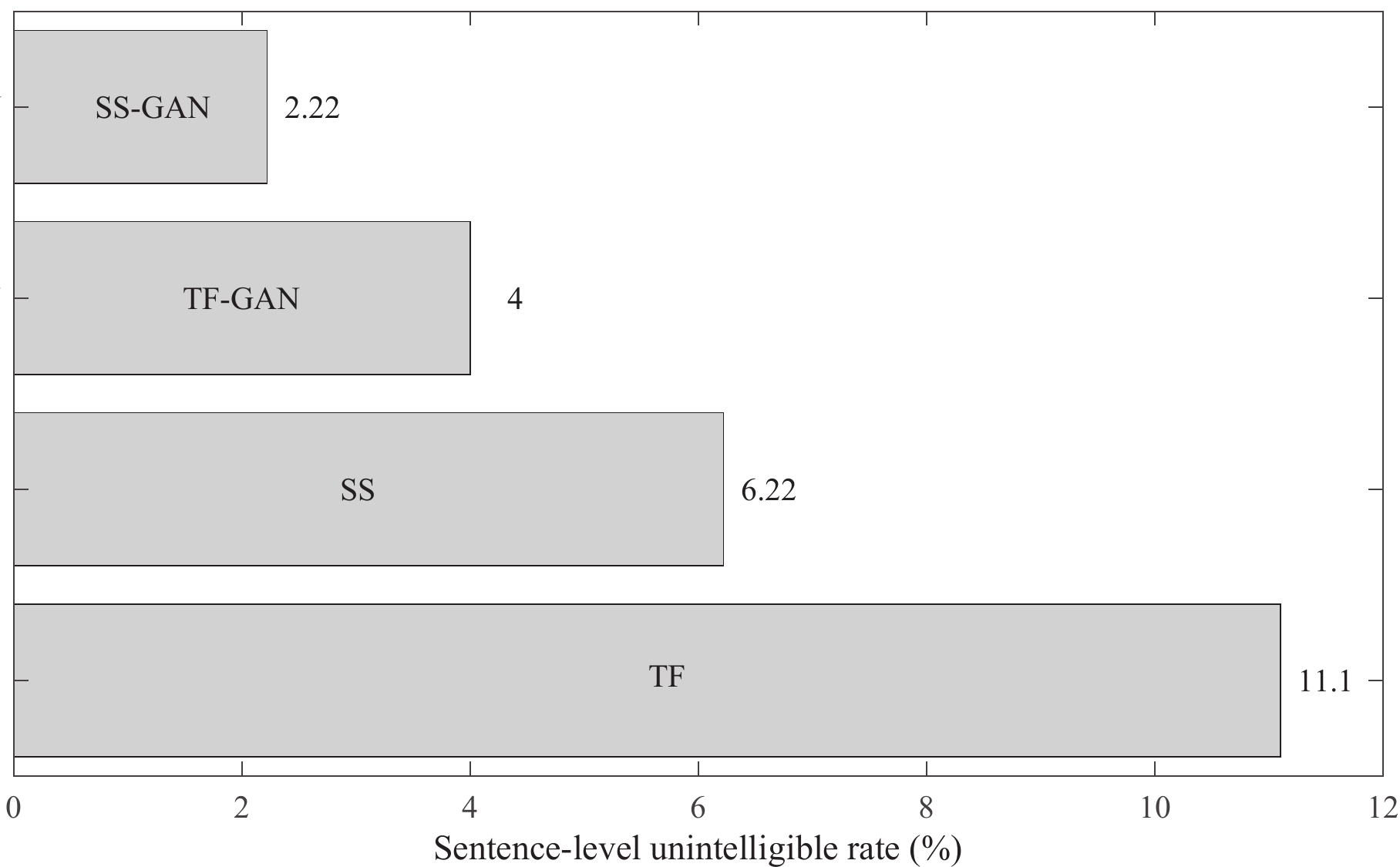}
  \caption{The results of the intelligibility tests}
  \label{fig:intelligibility}
\end{figure}

Table.\ref{tab:cmos} and Fig.\ref{fig:intelligibility} show the results of these two subjective evaluations. Compared with TF, both of CMOS score (-0.04) and preference (-2.2\%) show that the performance of SS is worse, but the intelligibility is improved on the pathological test set. In the comparison between TF-GAN and TF, the votes on TF-GAN is 5.33\% more than TF when 50\% of the votes are neutral. TF-GAN shows significantly better performance than TF with a higher CMOS (0.1) and preference (5.33\%). It also achieves a lower unintelligible rate (4\%) than TF (11.1\%) and SS (6.22\%). So compared with SS, GAN-based training algorithm is more effective. It can improve both naturalness and generalization for end-to-end TTS. As the combination of SS and GAN-based training algorithm, SS-GAN can further improve the intelligibility rate (2.22\%). SS-GAN does not achieve improvement in speech quality and naturalness due to SS, but has better performance in model generalization.

\subsection{Analysis}

\begin{figure}[htp]
  \centering
  \includegraphics[width=8cm]{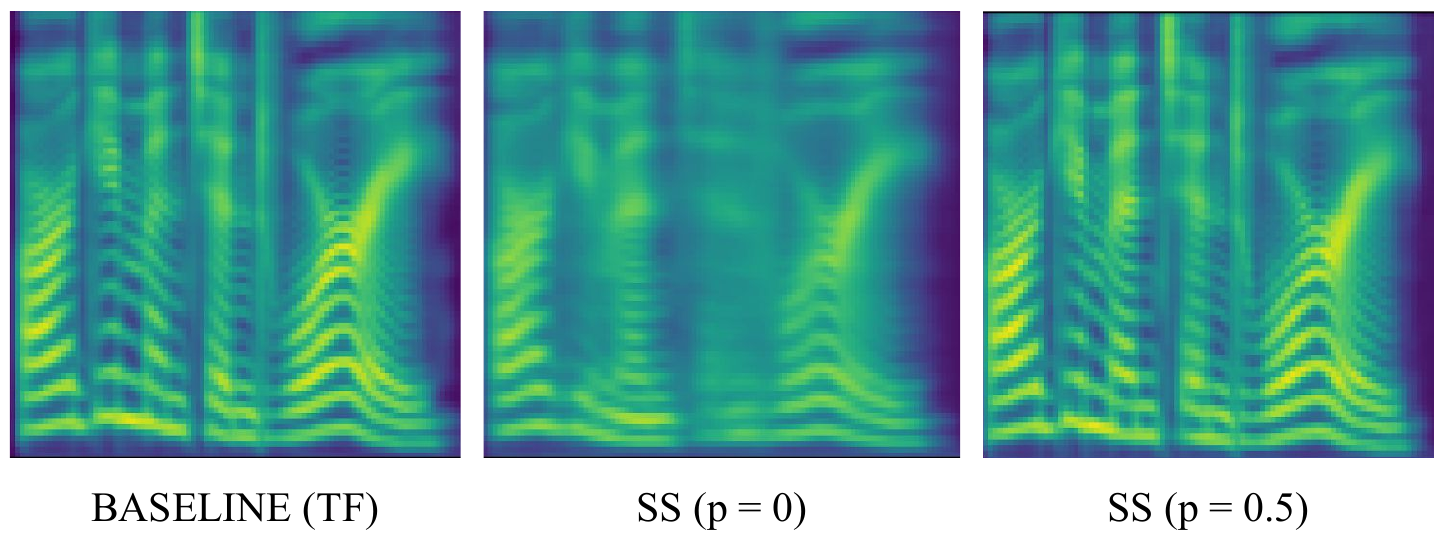}
  \caption{Mel spectrum synthesized by different models}
  \label{fig:sscase}
\end{figure}

We also try other decay strategies for scheduled sampling, but these experiments show that when we lower the sampling probability, more deterioration of the speech quality. Fig.\ref{fig:sscase} shows the Mel spectrum synthesized by the models trained with TF and SS. When we linearly decay the sampling probability from 1 to 0 within 50,000 iterations, the sound quality and clarity deteriorate significantly (shown in the middle part). It shows that calculating the frame-level loss for non-aligned data will lead to loss of model output quality, although predicted data is helpful for improving the generalization capability of TTS model. So we finally set the sampling probability to 0.5 to alleviate the misalignment between output and the target.

\begin{figure}[htp]
  \centering
  \includegraphics[width=8cm]{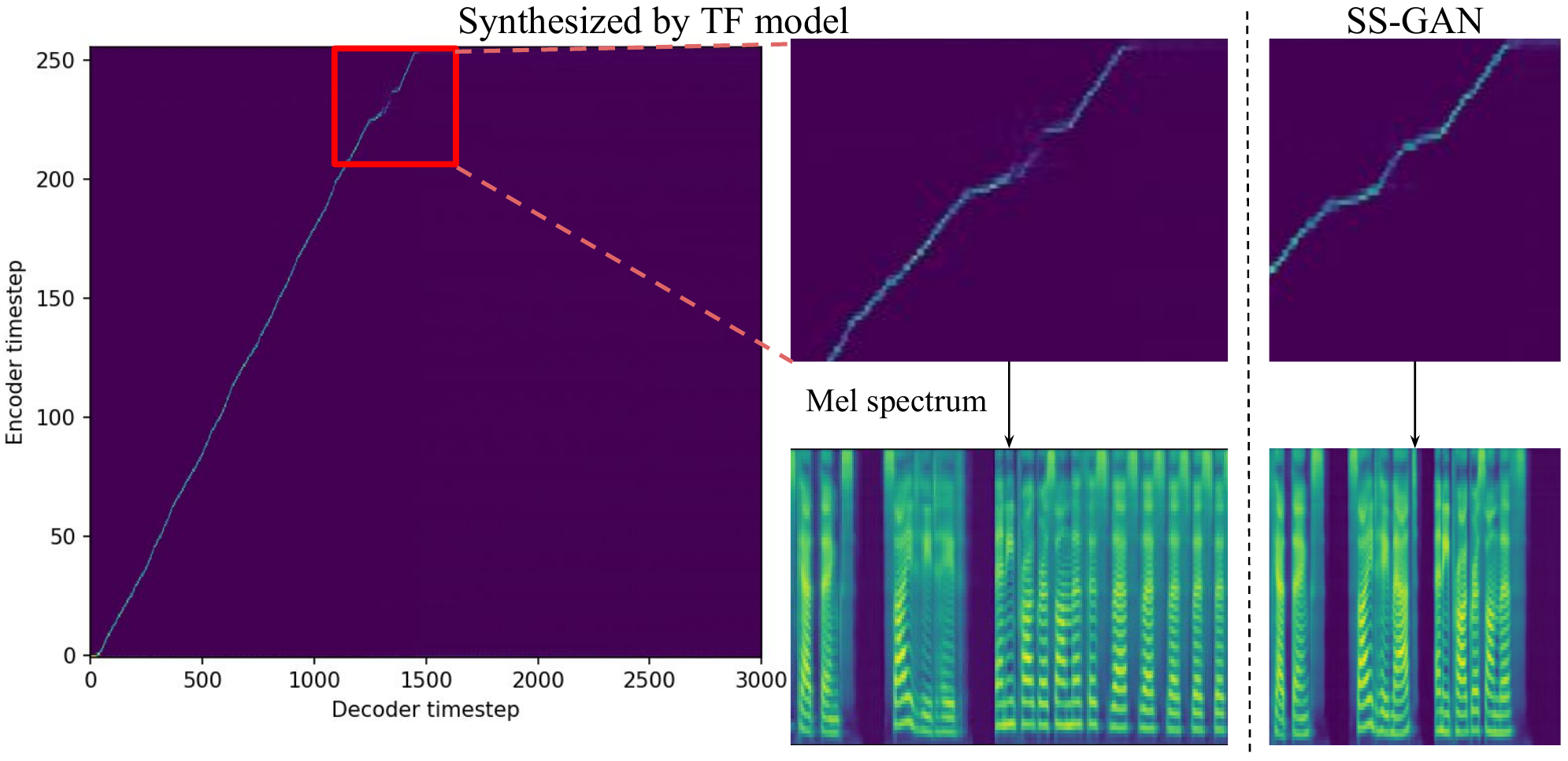}
  \caption{The alignment and Mel spectrum of a long sentence}
  \label{fig:badcase}
  \vspace{-0.3cm}
\end{figure}

To compare the performance before and after improving model generalization, we investigate the bad cases in the intelligibility test. We find that some long sentences can easily lead to garbled pronunciations, that is, the model suddenly starts to generate unintelligible and repeated garbled speech in decoding. Fig.\ref{fig:badcase} shows the alignment and Mel spectrum of such a case. We hypothesize this problem is due to the fact that a long and unseen context in the sentence can lead to higher cumulative errors in decoding. These errors, in turn, can distract the attention to the correct context. After improving the generalization with the proposed algorithm, the decoder is more robust in decoding noisy sequence. In our test set, over 50\% of these bad cases are fixed. The remaining bad cases with longer sentence and more complex contexts, may need better encoder to fix the problem.

\section{Conclusions}

This paper proposes a new GAN-based, end-to-end TTS training algorithm, which introduces the generated sequence to GAN training to avoid exposure bias in autoregressive decoder. Experimental results show that schedule sampling is harmful to synthesized speech quality, but can improve the model generalization capability of TTS model. Compared with scheduled sampling, our proposed algorithm improves both output quality and generalization of the model. By combining SS and GAN, we can further improve the generalization of the model by maintaining the speech quality and naturalness at the same level with a slight preference advantage of 3.8\%.

\clearpage

\bibliographystyle{IEEEtran}
\bibliography{mybib}


\end{document}